\documentclass[10pt,twocolumn,letterpaper]{article}

\usepackage[final]{cvpr}      

\usepackage{graphicx}
\usepackage{amsmath}
\usepackage{amssymb}
\usepackage{booktabs}
\usepackage{graphicx} 
\usepackage{booktabs}
\usepackage{multirow}
\usepackage[title]{appendix}
\usepackage{xspace}
\usepackage{mathtools}
\usepackage{bbding}
\usepackage{algorithm}
\usepackage{algorithmic}
\usepackage{graphicx}
\usepackage{amsmath}
\usepackage{amsthm}
\usepackage{booktabs}
\usepackage{verbatim}
\usepackage{multirow}
\usepackage{makecell}

\usepackage{pifont}
\usepackage{bbm}
\usepackage{url}
\usepackage{nicefrac}

\usepackage{colortbl}
\usepackage[table]{xcolor}
\definecolor{mygray}{gray}{.9}

\usepackage[pagebackref,breaklinks,colorlinks]{hyperref}

\usepackage[capitalize]{cleveref}
\crefname{section}{Sec.}{Secs.}
\Crefname{section}{Section}{Sections}
\Crefname{table}{Table}{Tables}
\crefname{table}{Tab.}{Tabs.}

\newcommand{\zbs}[1]{\textcolor{black}{#1}}

\usepackage[marginal]{footmisc}

\begin{document}

\title{Learning with Noisy labels via Self-supervised Adversarial Noisy Masking}

\author{Yuanpeng Tu$^{1}$\textsuperscript{*} \quad Boshen Zhang$^{2}$\textsuperscript{*} \quad Yuxi Li$^{2}$ \quad Liang Liu$^{2}$ \quad Jian Li$^{2}$ \\ Jiangning Zhang$^{2}$ \quad Yabiao Wang$^{2}$ \quad  Chengjie Wang$^{2}$ \quad Cai Rong Zhao$^{1}$\\
	$^{1}$Dept. of Electronic and Information Engineering, Tongji Univeristy, Shanghai \\ $^{2}$YouTu Lab, Tencent, Shanghai \\
	{\tt\small \{2030809, zhaocairong\}@tongji.edu.cn}\\
	{\tt\small \{boshenzhang, yukiyxli, leoneliu, swordli, vtzhang, caseywang, jasoncjwang\}@tencent.com}}

\maketitle

\footnote{$^{*}$ Yuanpeng Tu, Boshen Zhang contribute equally to this work.\\}

\begin{abstract}
   Collecting large-scale datasets is crucial for training deep models, annotating the data,  however, inevitably yields noisy labels, which poses challenges to deep learning algorithms. Previous efforts tend to mitigate this problem via identifying and removing noisy samples or correcting their labels according to the statistical properties (e.g., loss values) among training samples. In this paper, we aim to tackle this problem from a new perspective, delving into the deep feature maps, we empirically find that models trained with clean and mislabeled samples manifest distinguishable activation feature distributions. From this observation, a novel robust training approach termed adversarial noisy masking is proposed. The idea is to regularize deep features with a label quality guided masking scheme, which adaptively modulates the input data and label simultaneously, preventing the model to overfit noisy samples. Further, an auxiliary task is designed to reconstruct input data, it naturally provides noise-free self-supervised signals to reinforce the generalization ability of deep models. The proposed method is simple and flexible, it is tested on both synthetic and real-world noisy datasets, where significant improvements are achieved over previous state-of-the-art methods. 
\end{abstract}

\section{Introduction}
\label{sec:intro}
Deep learning has achieved remarkable success, relying on large-scale datasets with human-annotated accurate labels. However, collecting such high-quality labels is extremely time-consuming and expensive. As an alternative, inexpensive strategies are usually used for generating labels for large-scale samples, such as web crawling, leveraging search engines, or using machine-generated annotations. All these alternative methods inevitably yield numerous noisy samples. However, previous research~\cite{arpit2017closer} has revealed that deep networks can easily overfit to noisy labels and suffer from dramatic degradation in the generalization performance.

\begin{figure}[!t]
\centering
\includegraphics[scale=0.35]{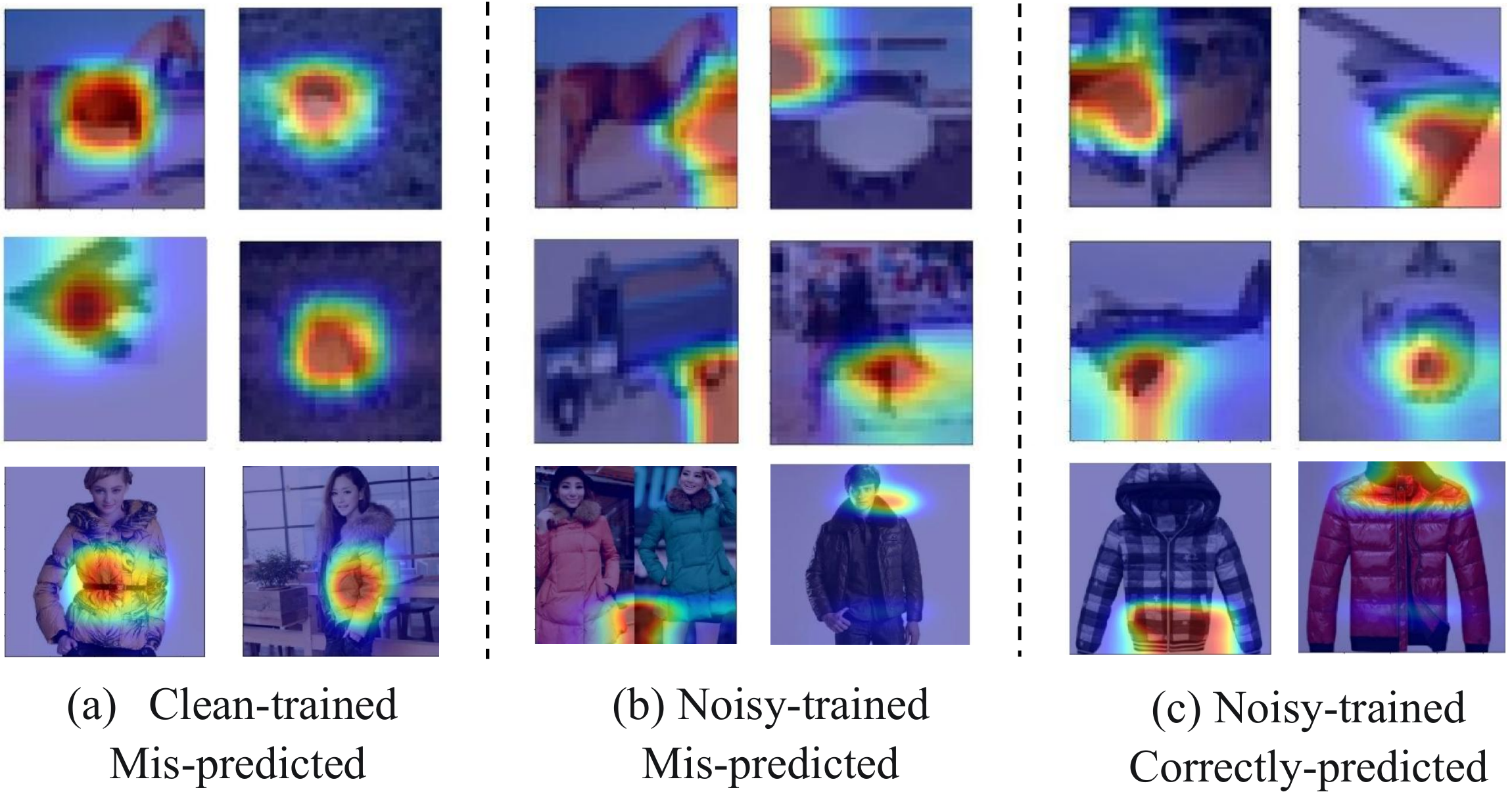}
\vspace{-0.6em}
\caption{Activation maps of mis-predicted (a-b) and correctly-predicted (c) samples when training PreAct ResNet-18 with clean (i.e., clean-trained) and noisy (i.e., noisy-trained) data on CIFAR-10 and Clothing1M~\cite{Clothing1M}.
}
\label{fig:motvation_v1}
\vspace{-1.0em}
\end{figure}

{Towards this problem, numerous Learning with Noisy labels (LNL) approaches {have been} proposed.}
{Sample selection methods~\cite{han2018co,yu2019does,wei2020combating,chen2019understanding} are the most straightforward methods, which} attempt to select clean samples based on certain criterion {(e.g., loss value)} and {then} reweight or {discard} the noisy instances to reduce the {{interference} of mislabeled training samples.}
However, these methods fail to {leverage} the potential information of the {discarded} samples. {Similarly}, label correction based methods~\cite{tanaka2018joint,yi2019probabilistic,zheng2021meta} try to correct labels, {which often impose assumptions on the existence of a small correctly-labeled validation subset or directly utilize predictions of deep models. However, such assumptions can not be always full-filled in real-world noisy datasets~\cite{song2019selfie}, leading to limited application scenarios. Besides, the predictions of deep model tend to fluctuate when training with mislabeled samples, making the label correction process unstable and sub-optimal~\cite{AAAI-2021-meta}.}
{On the contrary,} regularization based methods~\cite{arpit2017closer,song2022learning,liu2020early,zhou2021learning,ren2018learning,li2020dividemix,zheltonozhskii2022contrast,zhang2017mixup,goodfellow2014explaining} aim to alleviate the side effect of {label} noise by preventing the deep models from {overfitting to all training samples, which is flexible and can work collaboratively with other LNL methods}. {Both explicit and implicit regularization techniques were proposed, the}
former generally calibrates the parameter update of networks by modifying the expected training loss, \rm{i,e.}, dropout and weight decay. {The latter} focuses on improving the generalization by utilizing the stochasticity, \rm{i,e.}, data augmentation strategy and stochastic gradient descent~\cite{song2022learning}. 
{However, most of these regularization methods are designed for general fully-supervised learning tasks with correctly-labeled training samples, and poor generalization ability could be obtained when the label noise is severe~\cite{song2022learning}}.

\begin{figure*}[!t]
 \centering
\small
 \begin{minipage}{1.0\textwidth}
    \centering
    \includegraphics[width=\textwidth]{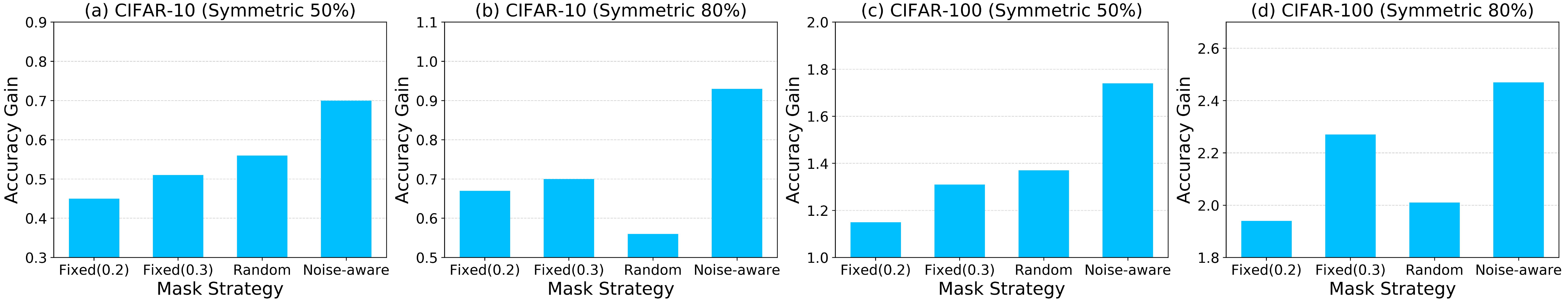}
\end{minipage}
\vspace{-2mm}
  \caption
    {
    \small
        {A experiment for masking the max-activated region with different mask ratios. The performance gains of different mask strategies under 50\% and 80\% symmetric noise of CIFAR-10/100~\cite{datasetcifar10} are reported}, where DivideMix~\cite{li2020dividemix} is adopted as the baseline. "Fixed(0.2/0.3)" denotes masking all the images with the same mask ratio of 0.2/0.3. "Random" represents masking images with a random mask ratio between 0.2 and 0.4. "Noise-aware" is masking noisy samples with a mask ratio of 0.3 while the ratio for clean ones is 0.2.   
      }
  \label{fig:motivation_work}
  \vspace{-4mm}
 \end{figure*}

In this paper, {we present a novel regularization based method specialized for LNL problem. Our method is inspired by an observation that some differences exist} in activation maps between the predictions of deep models trained with clean and noisy labels.
As shown in Fig.~\ref{fig:motvation_v1}~(a), activation maps of {mis-predicted} samples by the model trained with clean labels are always focused on their foreground areas. By contrast, {when} the model is trained with noisy labels, {it tends to} generates results focused on the meaningless background area {for the mis-predicted samples~(Fig.~\ref{fig:motvation_v1}~(b)). And even for cases that the prediction is correct, i.e., Fig.~\ref{fig:motvation_v1}~(c), the high-activated region drifts to irrelevant (i.e., edge) areas of the object, rather than the {main regions of the object}. This indicates that \emph{model trained with mislabeled samples is likely to overfit {to 
some corner parts of the object from} noisy data {or remember} the less-informative regions (i.e., background)}.}
Therefore, {from this observation}, we propose a novel Self-supervised Adversarial Noisy Masking (SANM) method for LNL. {The idea is to explicitly regularize the model to generate more diverse activation maps through masking the certain region (e.g., max-activated) of the input image}
and thus alleviates the confirmation bias~\cite{han2018co}.

Moreover, {to investigate how the masking strategies affect the performance of training deep models with noisy labels. We conduct a simple experiment, where different adversarial masking strategies (i.e., masking the max-activated region of input images with fixed, random, and noise-aware mask ratio) were applied to the popular LNL framework DivideMix~\cite{li2020dividemix}.} As shown in Fig.~\ref{fig:motivation_work}, we find that the performance {gain} is much more sensitive to the masking strategy. {A fixed or random mask ratio obtain only marginal gain, but when dealing with the correctly-labeled and mislabeled samples with different mask ratios, a significant performance gain is achieved. This indicates that how to design an effective adversarial masking method in the case of LNL is non-trivial and not well addressed.}


{Towards the problem above, a label quality guided adaptive masking strategy is proposed,} which modulates the {image label and the mask ratio of image masking simultaneously}. {The label is updated via first leveraging the soft-distributed model prediction and then reducing the probability of max-activated class with the noise-aware masking ratio, while at the same time lifting the probability of other classes.}
Intuitively, a sample with a high probability of being mislabeled will possess a larger masking ratio, leading to a more strictly regularized input image and modulated label. Therefore, the negative impact of noisy labels can be greatly reduced. As for the correctly-labeled samples, the masking ratio is relatively small, which plays the role of general regularization strategy and improves the model generalization ability by preventing the model from overfitting training samples. Further, we {specially customized} an auxiliary decode branch to reconstruct the original input image, which provides noise-free self-supervised information for learning a robust deep model. The proposed SANM is flexible and can further boost existing LNL methods. It is tested on synthetic and real-world large-scale noisy datasets, and elaborately ablative experiments were conducted to verify each design component of the proposed method. In a nutshell, the key contributions of this work are:

$\bullet$ We propose a novel self-supervised adversarial noisy masking method named SANM to explicitly impose regularization {for LNL problem}, preventing the model from overfitting to {some corner parts of the object or less-informative regions from noisy data;}

$\bullet$ A label quality guided masking strategy is proposed to {differently adjust} the process for clean and noisy samples according to the label quality estimation{. This strategy}
modulates {the image label and the ratio of image masking} simultaneously;


$\bullet$ A self-supervised mask reconstruction auxiliary {task} is designed to reconstruct the original images based on the features of masked ones, which aims at enhancing generalization by providing noise-free {supervision} signals.

The source code will be published online to reproduce our experimental results.

\section{Related Work}
\subsection{Learning with Noisy Labels}

A variety of methods have been proposed to improve the robustness of DNNs on noisy datasets, which consist of regularization, label correction, and noisy sample selection. 

\textbf{Regularization} based methods attempt to help resist memorizing noisy data since previous works ~\cite{arpit2017closer} have revealed that networks always firstly learn simple patterns before memorizing noisy hard data. Previous research~\cite{song2022learning} claims that these methods can be roughly divided into two categories: explicit regularization and implicit regularization. Explicit regularization based methods generally explicitly modify the losses for the training samples. Among them, ELR~\cite{liu2020early} tries to alleviate the influence of noisy data by regularizing the loss gradient with estimated target probabilities. SR~\cite{zhou2021learning} proposes a spare regularization method to approximately restrict network output to the permutation set over a one-hot vector. Meta learning based methods~\cite{ren2018learning} impose regularization on the sample gradient directions by bi-level optimization with the assistance of a small unbiased validation. In addition to these regularizers, example re-weighting strategies are also popular explicit regularization solutions to the over-fitting problem. However, these methods often need sensitive hyper-parameters or deeper networks to improve the model capacity. For implicit regularization methods, they generally utilize the stochasticity effect, i.e., augmentation and stochastic gradient descent. Among them, semi-supervised based LNL methods~\cite{li2020dividemix, zheltonozhskii2022contrast} generally use mixup~\cite{zhang2017mixup} as an effective regularization strategy to hinder over-fitting, which makes the network favor simple linear behavior in-between training examples. Adversarial samples based methods~\cite{goodfellow2014explaining} improve the robustness by helping the model classify original and perturbed inputs. Since the model is unable to distinguish right from wrong by itself, regularization can generally work surprisingly well in certain synthetic noisy cases. 

\textbf{Label correction} methods seek to transform noisy labels to correct ones directly. The pseudo labeling strategy is commonly adopted by these methods, which utilizes the confident output of the network. Joint~\cite{tanaka2018joint} optimizes labels and network parameters in an alternating strategy. PENCIL~\cite{yi2019probabilistic} corrects and updates labels probabilistically through the back-propagation mechanism. MLC~\cite{zheng2021meta} trains a label correction network through meta learning to generate pseudo labels for the training of main models. However, these approaches either need extra clean validation data or suffer from significant performance drop under high noise settings due to the low quality of generated pseudo labels.

\textbf{Noisy sample selection} approaches mostly exploit the small-loss criterion, which is based on the observation that samples with noisy labels tend to have much higher losses than the clean ones. Among these methods, MentorNet uses a pre-trained teacher model to select clean samples for the training of a student model. Co-teaching~\cite{han2018co} and Co-teaching+~\cite{yu2019does} update the parameters of two parallel networks with clean instances chosen from their peer networks, which target avoiding the confirmation bias problem. JoCoR~\cite{wei2020combating}, while INCV~\cite{chen2019understanding} attempts to separate clean data from the noisy training set by using cross-validation and removing high-loss samples. Though these methods may reduce the risk of error correction, they fail to use the potential information of the abandoned samples. Moreover, most of them need prior knowledge, i.e., the noise rate.

\subsection{Masking Regularization Techniques}

As a commonly {adopted} strategy in weakly-supervised semantic segmentation, adversarial masking regularization aims at helping models discover {less apparent} semantic objects. Specifically, a {Class Activation feature Map (CAM)~\cite{gradCAM}} branch is first used to generate initial attention maps and selectively mask the discovered areas in the images through a threshold. Then the masked images are used for the training of another network. AE-PSL~\cite{wei2017object} iteratively erases the images and expands target object regions with the erased areas to produce final contiguous segmentation masks. Self-erasing~\cite{hou2018self} utilizes the background prior to make models generate more refined and focused attention on the foreground area. GAIN~\cite{li2018tell} tries to impose constraints on the activation maps in a regularized style to generate more complete attention regions. PSA~\cite{stammes2021find} trains two networks with opposing losses to eliminate the drawbacks of certain mask strategies. In addition to weakly-supervised semantic segmentation, adversarial masking is also used in fine-grained visual classification. DB~\cite{sun2020fine} designs a diversification block to mask out the most salient features, aiming at forcing networks to mine subtle differences between images. WS-DAN~\cite{hu2019see} randomly crop and drop one of the discriminative regions in the attention maps to encourage models to mine local features. \textcolor[rgb]{0,0,0}{Several adversarial masking strategies are proposed in ~\cite{gupta2021adversarial,yang2021towards} to preserve human privacy by masking the minimal number of pixels required for identification. ~\cite{valvano2021re} designs an adversarial discriminator for mask generation at inference to correct segmentation errors.} However, the purpose of these methods is quite different from our approach. Based on the {assumption} that the samples are all correctly-labeled, previous methods generally retain the original most salient regions while helping models mine more subtle information through masking regularization. By contrast, our method aims to regularize the model through adversarial masking to make the activation map more dispersed rather than concentrated to specific regions, {and a similar masking strategy is applied to generate more smooth pseudo labels}, thus avoiding over-fitting to noisy labels and alleviating the confirmation bias. 

\label{headings}

\begin{figure*}[!t]
\centering
\includegraphics[width=1.0\textwidth]{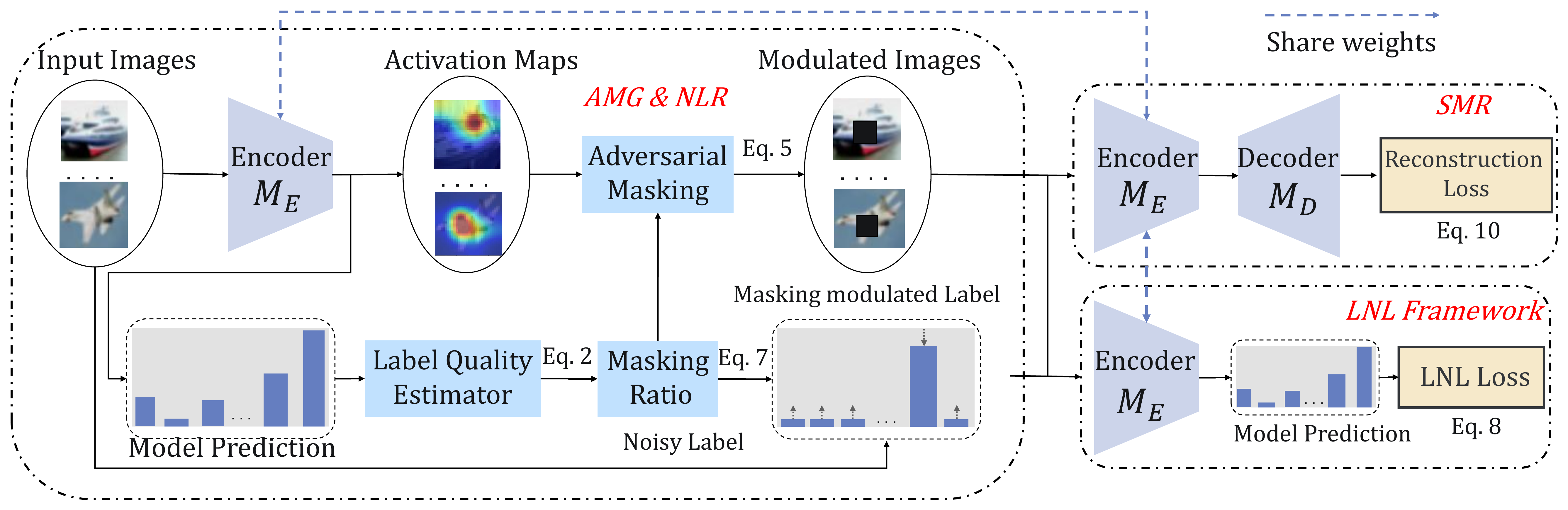}
\vspace{-2em}
\caption{{The technical workflow of the proposed SANM. Three components are included in SANM: AMG (Adversarial Masking Generation), NLR (Noisy Label Regularization), SMR (Self-supervised Masking Reconstruction). In AMG and NLR, firstly feed the images to an encoder and generate activation maps. And a label quality guided adversarial masking strategy is proposed to modulate the images and noisy labels simultaneously. Further, an auxiliary decode branch is designed in SMR to reconstruct input images from the features of masked images. Finally, the generated modulated images and labels of SANM together with the reconstruction loss can be directly adopted for the training of existing LNL framework.}}
\label{fig:concept}
\vspace{-1em}
\end{figure*}

\subsection{Overview (Framework)}

For convenience, we begin by introducing the notations. Consider LNL as a classification task, the class number is $c$. The noisily-labeled training dataset is denoted as $ D = \{{(x_i, y_i)}\}_{i=1} ^{N}$, where $x_i \in \mathbb{R}^{H_i\times W_i\times 3}$ and $y_i \in \{0, 1\}^c$ are the $i$-th sample and its noisy label, $H_i, W_i$ are the height and width of samples. $N$ represents the dataset size. The encoder and decoder models are denoted as $M_E$ and $M_D$ respectively. We firstly introduce the overall framework of the proposed SANM. As shown in Fig.~\ref{fig:concept}, a label quality guided masking scheme is proposed to adaptively perform adversarial noisy masking for each sample where a contrastive regularization strategy is imposed on the noisy predictions as well. Afterward, the representations of the masked images are used for the self-supervised reconstruction of original images, which aims at preventing models from overfitting to the noisy labels by providing noise-free signals. For application, SANM can be easily integrated with existing LNL frameworks to further boost their performance.

\subsection{Adversarial Noisy Masking}

\textbf{Adversarial Masking Generation.} To highlight the semantic regions of correctly labeled samples and meanwhile keep the captured attention areas of the noisy samples from expanding to meaningless background regions, we propose a novel \textbf{A}dversarial noisy \textbf{M}asking \textbf{G}eneration (AMG) strategy during the training process. Specifically, we begin by extracting the feature of image $x_i$ through $M_E$ and the obtained feature map and prediction are denoted as $F_i \in \mathbb{R}^{H_F\times W_F\times N_C}$ and $\widetilde{y_i}$ respectively, where $H_F, W_F, N_C$ are the height, width and channel number of the feature layer. For each sample, a two-component Gaussian Mixture Model (GMM)~\cite{permuter2006study} is fit to the cross-entropy loss with the Expectation-Maximization algorithm, where the generated weight for $x_i$ is denoted as $G_i$. Afterward, $F_i$ is used to generate the activation maps~\cite{gradCAM} $A_i \in \mathbb{R}^{H_i\times W_i}$ of the category with the largest prediction, which can be derived as:
\begin{equation}
    A_i = \text{CAM}(F_i, \enspace \text{argmax}(\widetilde{y_i}))
    \label{eq:activation}
\end{equation}
where $\text{CAM}(\cdot)$ is the activation map generation function. Then the coordinates with the maximum and minimum activation values for $x_i$ are recorded as $\{{h}_i ^{max}, {w}_i ^{max}\}$ and $\{{h}_i ^{min}, {w}_i ^{min}\}$ respectively. For the $\{{h}_i ^{max}, {w}_i ^{max}\}$, assuming that the GMM division is overall correct, it is believed that the activation maps of correct-labeled samples are mainly focused on the target area of the foreground, while due to the interference of noisy labels, the max-activated region in the activation maps of mislabeled samples are mostly concentrated in meaningless background areas. 
{To breakdown the over-fitting to noisy labels, we intend to adversarially augment training samples via a novel making strategy. In particular, denote the loss function as $\mathcal{L}(x_i,y_i)$, we try to maximize $\mathcal{L}$ by adaptively generating region perturbations for noisy and clean samples across the training process. The objective can be derived as $\max \mathcal{L}(x_i+\sigma, y_i)$, where $\sigma$ is the generated perturbations. }
Specifically, we generate masks with larger ratios for the max-activated region of noisy samples to force the model to reduce its response to these areas and pay more attention to the rest area, where the target is more likely to be found. For the correct-labeled ones, high response areas in their activation maps are likely to exist in the edge area of the object since they are affected by the gradient of noisy samples. Thus smaller mask ratios are used for the max-activated regions of them, which keep the model from ignoring the masked area and force it to mine discriminative information from the rest areas of the target since the correct labels help limit the network to pay more attention to the foreground area. However, it will make the activation map of the well-trained model scattered to the meaningless background when filtering out nearly all the peak locations for each sample through a threshold like existing methods, which violates our purpose. Therefore, we choose to mask random rectangular areas based on recorded coordinates. Specifically, the mask ratio $r_i$ for $x_i$ is calculated as follows:
\begin{equation}
    r_i = \mu\times (1 - G_i)
\label{eq:ratio}
\end{equation}
where $\mu$ is the basic mask ratio. {The more likely a sample is to be noisy, the smaller the $G_i$ is. Based on Eq.~(\ref{eq:ratio}), the greater mask ratio will be adopted for this sample, thus stronger regularization will be imposed~\cite{li2020dividemix}}. Subsequently, the aspect of the masked area $\delta_i$ is generated as follows:
\begin{equation}
    \delta_i \sim \text{Uniform}(\delta, \enspace \frac{1}{\delta})
\end{equation}
where $\delta$ is a hyper-parameter for generating the aspect. Based on $r_i$ and $\delta_i$, the height and width of the masked area are given by:
\begin{equation}
\left\{\begin{array}{c}
h_{i}^{u p}=\max \left(h_{i}^{\max }-\sqrt{\frac{\left(H_{x} \times W_{x}\right) \times r_{i} \times \delta_{i}}{4}}, 0\right) \\
h_{i}^{\text {dn }}=\min \left(h_{i}^{\max }+\sqrt{\frac{\left(H_{x} \times W_{x}\right) \times r_{i} \times \delta_{i}}{4}}, H_{x}\right) \\
w_{i}^{\text {lt }}=\max \left(w_{i}^{\max }-\sqrt{\frac{\left(H_{x} \times W_{x}\right) \times r_{i}}{4 \delta_{i}}}, 0\right) \\
w_{i}^{\text {rt }}=\min \left(w_{i}^{\max }+\sqrt{\frac{\left(H_{x} \times W_{x}\right) \times r_{i}}{4 \delta_{i}}}, W_{x}\right)
\end{array}\right.
\end{equation}
The selected locations in the masked area of $x_i$ are assigned random values generated from a uniform distribution. Therefore, the adversarial noisy masked image $x_i ^{m}$ can be derived as follows:
\begin{equation}
    x_{i}^{m}(m, n)=\left\{\begin{array}{l}
 {U}(0,1), \text { if } m \in\left[h_{i}^{up}, h_{i}^{dn }\right], n \in\left[w_{i}^{lt }, w_{i}^{rt}\right], \\
x_{i}(m, n), \quad \quad  \text { otherwise. }
\end{array}\right.
\end{equation}
where $U$ denotes the Uniform distribution. Similar operation is performed on the regions of minimum activation value with the coordinates $\{{h}_i ^{min}, {w}_i ^{min}\}$ as well. Different from Eq.~(\ref{eq:ratio}), since the regions of minimum activation value for correct-labeled samples are mainly focused on the background, a smaller mask ratio is adopted for them, helping models narrow the search scope for targets. Therefore, we set the mask ratio of the regions with minimum activation values as $\mu\times G_i$.

\textbf{Noisy {Label} Regularization.} To penalize the {over-confident} predictions of the network, a contrastive \textbf{N}oisy \textbf{L}abel \textbf{R}egularization (NLR) strategy is designed. Specifically, the generated $x_i ^{m}$ is fed into $M_E$ to obtain the prediction $\widetilde{{y}_i ^m}$ and feature $\mathbf{f_i} ^{m}$ as follows:
\begin{equation}
{\mathbf{f_i} ^{m}}=M_E({x_i ^{m}};\theta_{E})
\label{eq:feature}
\end{equation}
where $\theta_{E}$ denotes the parameters of the encoder $M_E$. Specifically, we give the pseudo labels of $x_i ^{m}$ with the predictions of $x_i$. Since the mask generation is based on the activation maps of the class with the largest predictions, the corresponding prediction is penalized with the mask ratio $r_i$. Besides, there exists an equal possibility for the masked region to be predicted as each class. Thus the prediction values for all the classes are increased uniformly. To sum up, the regularized pseudo label ${{y}_i ^r}$ for $x_i ^m$ can be derived as follows:
{\begin{equation}
{{y}_{i} ^{r}}(j)=\left\{\begin{array}{l}
{y}_{i}(j)-r_{i}+r_{i} / c, \text { if } j=\operatorname{argmax}\left({y_{i}}(j)\right) \\
{y}_{i}(j)+r_{i} / c, \quad \quad  \text { otherwise }
\end{array},\right.
\end{equation}}where $0 \leq j <c$. Afterward, ${{y}_i ^r}$ and $\widetilde{{y}_i ^m}$ are used for supervised loss calculation as:
{\begin{equation}
   {\mathcal{L}_{\rm c}}={\mathcal{L}_{\rm ce}}({\widetilde{{y}_i ^m}},{{{y}_i ^r}})
\end{equation}}
where $\mathcal{L}_{\rm ce}$ denotes the cross-entropy loss.
\vspace{-2mm}
\subsection{Self-supervised Masking Reconstruction}
Inspired by~\cite{he2021masked}, a \textbf{S}elf-supervised \textbf{M}asking \textbf{R}econstruction branch (SMR) is designed to impose implicit regularization by providing extra noise-free supervised signal. Specifically, the decoder $M_D$ reconstructs the original image from the latent representation of the masked image ${x}_i ^m$ by predicting the pixel values for the masked patch, which is defined as:
\begin{equation}
{{x_i} ^{r}}=M_D({\mathbf{f_i} ^{m}};\theta_{D})
\label{eq:decode}
\end{equation}
where $\theta_{D}$ is the parameters of the decoder model $M_D$. Then the mean squared error (MSE) loss is computed between the reconstructed image ${x}_i ^r$ and the original unmasked image $x_i$ in the pixel space, which is shown as follows:
\begin{equation}
   {\mathcal{L}_{\rm r}}=\|{{x}_i ^r}-{{x}_i}\|^{2}
\end{equation}
In this way, a challenging self-supervised pretext task that requires holistic understanding beyond low-level image statistics is created for the model. {This supervision signal is not only free of interference from noisy labels but can also enhance the model generalization ability.}
Therefore the overall loss function for SANM is constructed as follows:
\begin{equation}
   {\mathcal{L}_{\rm train}}={\mathcal{L}_{\rm c}} + \beta{\mathcal{L}_{\rm r}}
   \vspace{-6mm}
\end{equation}
\begin{table*}[!t]
\footnotesize
    \centering
    \caption{
        Comparison with state-of-the-art methods on CIFAR-10/100 datasets with symmetric noise.
        }
    \vspace{-1em}
    \setlength\tabcolsep{10pt} \renewcommand{\arraystretch}{1.3}
    \resizebox{0.83\textwidth}{!}{$
    \begin{tabular}{l |c  c  c c| c c cc}
        \hline
        Dataset                                                                &      &                  \multicolumn{3}{c|}{CIFAR-10}                  &                    \multicolumn{4}{c}{CIFAR-100}                     \\  
        Method/Noise                                                        ratio      & 20\%             &     50\%      &     80\%      &     90\%      &       20\%       &       50\%       &     80\%      &     90\%      \\ \hline
       {Cross-Entropy (CE)}                                          & 86.8             &     79.4      &     62.9      &     42.7      &       62.0       &       46.7       &     19.9      &     10.1      \\
                                                                                \hline
        {Co-teaching$+$~\cite{yu2019does}}                        & 89.5             &     85.7      &     67.4      &     47.9      &       65.6       &       51.8       &     27.9      &     13.7      \\
                                                                                \hline
        {Mixup~\cite{zhang2017mixup}}                            & 95.6             &     87.1      &     71.6      &     52.2      &       67.8       &       57.3       &     30.8      &     14.6      \\
                                                                                \hline
        {PENCIL~\cite{yi2019probabilistic}}                      & 92.4             &     89.1      &     77.5      &     58.9      &       69.4       &       57.5       &     31.1      &     15.3      \\
                                                                                \hline
        {Meta-Learning~\cite{MLNT}}                              & 92.9             &     89.3      &     77.4      &     58.7      &       68.5       &       59.2       &     42.4      &     19.5      \\
                                                                               \hline
        {M-correction~\cite{arazo2019unsupervised}}              & 94.0             &     92.0      &     86.8      &     69.1      &       73.9       &       66.1       &     48.2      &     24.3      \\
                                                                                \hline
        {DivideMix~\cite{li2020dividemix}}             & {96.1} &     {94.6}      &     {93.2}      &     76.0     &       {77.3}       &       {74.6}       &     60.2      &     31.5      \\
                                                                                \hline
        {C2D~\cite{zheltonozhskii2022contrast}}                         & {96.3} & 95.2 & 94.4 & 93.5 & 78.6 & 76.4 & 67.7 & {58.7} \\  \hline                                                                            
        {AugDesc~\cite{nishi2021augmentation}}                         & 96.3 & 95.4 & 93.8 & 91.9 & 79.5 & 77.2 & 66.4 & 41.2 \\
                                                                               \hline                                                                       
        {GCE~\cite{ghosh2021contrastive}}                         & 90.0 & 89.3 & 73.9 & 36.5 & 68.1 & 53.3& 22.1 & {8.9} \\
                                                                              \hline
                                                                               
        {Sel-CL+~\cite{li2022selective}}                         & 95.5 & 93.9 & 89.2& 81.9 & 76.5 & 72.4 & 59.6 & {48.8} \\
                                                                               \hline

       {MOIT+~\cite{ortego2021multi}}                         & 94.1 & 91.8 & 81.1 & 74.7 & 75.9 & 70.6 & 47.6 & {41.8} \\
                                                                                 \hline \hline 
        {SANM(DivideMix)}                             & {96.4} & {95.8} & {94.6} & {92.3}  & {81.2} & {78.2} & {68.7} & {43.5}  \\  \hline                                                             
                  
        {SANM(C2D)}                                        & \textbf{96.6} & \textbf{96.4} & \textbf{95.7} & \textbf{95.1}  & \textbf{81.9} & \textbf{79.3} & \textbf{71.6} & \textbf{61.9}  \\  
                                                                     \hline
    \end{tabular}$}
    \label{tab:cifar_sym}
    \vspace{-2mm}
\end{table*}
\\
where $\beta$ is a parameter to balance the weight of two loss terms. Putting this all together, Algorithm \ref{pseudoalgorithm} delineates the proposed SANM. Firstly, the original images are masked with the label quality guided masking scheme, where mislabeled and correct-labeled samples are modulated adaptively. Meanwhile, the predictions of the original images are used for regularized label generation. Then the regularized features of the masked images are utilized for self-supervised masking reconstruction, which provides noise-free supervisions. Finally, SANM can be plugged into existing LNL frameworks to further boost performance.

\section{Experiments}
\vspace{-1em}

\begin{algorithm}[!t] 
    \footnotesize
    \caption{The proposed SANM framework}
    \label{alg:dmlp}
    \begin{algorithmic}[1]
    \renewcommand{\algorithmicrequire}{\textbf{Input:}}
    \REQUIRE Noisy training set $D$, encoder model $M_E(\cdot;\theta_{E})$, decoder model $M_D(\cdot;\theta_{D})$, batch size $b$, max iterations $m$, basic mask ratio $\mu$.
    \renewcommand{\algorithmicrequire}{ \textbf{Procedure:}}
    \REQUIRE
    \FOR{$i = 1$ to $m$}
    \STATE /*\textbf{AMG} starts*/
    \STATE $\{x_i, y_i\}_{i=1} ^{b}$ $\leftarrow$ SampleMiniBatch($D, b$).
    \STATE Feed $\{x_i\}_{i=1} ^{b}$ into $M_E$ and generate feature maps $\{F_i\}_{i=1} ^{b}$ and predictions $\{\widetilde{y_i}\}_{i=1} ^{b}$.
    \STATE Generate activation maps $\{A_i\}_{i=1} ^{b}$ by Eq.~(\ref{eq:activation}).
    \STATE Calculate mask ratios $\{r_i\}_{i=1} ^{b}$ and adversarial masked images $\{x_i ^{m}\}_{i=1} ^{b}$ by Eq.~(2-5).
    \STATE /*\textbf{NLR} starts*/
    \STATE Feed $\{x_i ^{m}\}_{i=1} ^{b}$ into $M_E$ and generate predictions $\{{\widetilde{{y}_i ^m}}\}_{i=1} ^{b}$ and features $\{\mathbf{f}_i ^{m}\}_{i=1} ^{b}$ by Eq.~(6).
    \STATE Calculate the regularized labels $\{{{y}_{i} ^{r}}\}_{i=1} ^{b}$by Eq.~(7).
    \STATE Calculate cross-entropy loss $\mathcal{L}_c$ by Eq.~(8).
    \STATE /*\textbf{SMR} starts*/
    \STATE Feed $\{\mathbf{f}_i ^{m}\}_{i=1} ^{b}$ into $M_D$ and generate reconstructed images $\{x_i ^r\}_{i=1} ^{b}$ by Eq.~(9).
    \STATE Calculate self-supervised reconstruction loss $\mathcal{L}_r$ by Eq.~(10).
    \STATE Update parameter $\theta_E, \theta_D$ in backward process.
    \ENDFOR
    \renewcommand{\algorithmicensure}{\textbf{Output:}}
    \ENSURE The final encoder $M_E(\cdot;\theta_E)$.
    \end{algorithmic}  
    \label{pseudoalgorithm}
     
\end{algorithm}

\subsection{Datasets and Implementation Details}
\textbf{Simulated noisy datasets.} We evaluate on two simulated noisy datasets: CIFAR-10/100~\cite{datasetcifar10}, which consists of 50k training samples and 10k test samples. Following previous work~\cite{li2020dividemix}, two simulated noisy settings are considered: symmetric and asymmetric noise. 
Multiple noise rates are considered: $\tau \in \{20\%, 50\%, 80\%, 90\%\}$ for symmetric noise and $\tau \in \{20\%, 40\%\}$ for asymmetric noise. An 18-layer PreAct ResNet is used and we train for 300 epochs with a batch size of 64 and SGD optimizer. The initial learning rate is set as 0.02 and is reduced by a factor of 10 after 150 epochs. $\beta$ is empirically set as 1.

\textbf{Real-world noisy datasets.} We further validate our method on large-scale real-world noisy dataset, i.e.,, Clothing-1M~\cite{Clothing1M} and Animal-10N~\cite{animal_SELFIE}, the former contains 1 million training samples with nearly 40\% mislabeled samples, the latter contains 55,000 human-labeled online images for 10 confusing animals. It includes approximately 8\% noisy-labeled samples.
For a fair comparison, the original protocol is used to split the training and test images. We strictly follow the previous experimental settings, more details can be found in the supplementary materials. $\beta$ is empirically set as 1.

\begin{figure*}[!t]
 \centering
\small
 \begin{minipage}{0.9\textwidth}
    \centering
    \includegraphics[width=\textwidth]{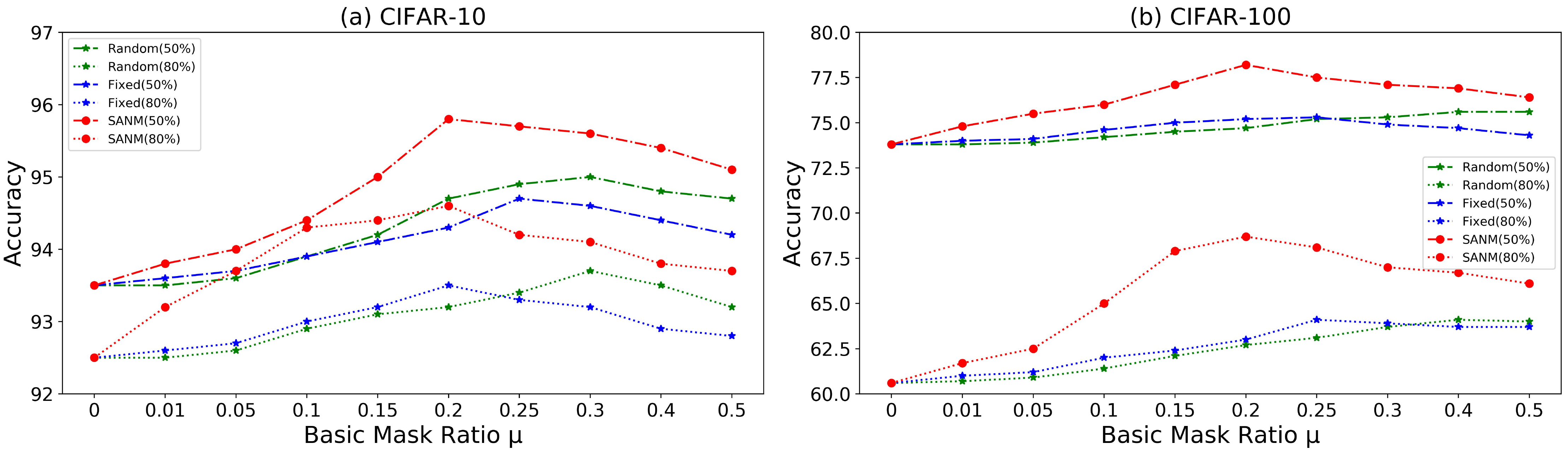}
\end{minipage}
 \vspace{-1em}
  \caption
    {
    \small
        Investigation of the basic mask ratio $\mu$ under 50\% and 80\% symmetric noise of CIFAR-10/100.
      }
  \label{fig:ablation_parameter}
 \vspace{-2mm}
 \end{figure*}

\begin{table*}[t]
\begin{minipage}{0.66\linewidth}
\begin{minipage}{0.47\linewidth}
    \footnotesize
    \caption{Asymmetric noise on CIFAR-10.}
    \vspace{-1em}
    \centering
    \renewcommand{\arraystretch}{1.1}
    {\begin{tabular}{p{2.4cm}|cc}
        \hline
        \multirow{2}{*}{Method}  & \multicolumn{2}{c}{Noisy ratio} \\ \cline{2-3} 
        & 20\% & 40\% \\ \hline
        Joint-Optim\cite{tanaka2018joint}              & 92.8    & 91.7 \\ 
        
        PENCIL~\cite{yi2019probabilistic}                & 92.4   & 91.2 \\ 
        F-correction~\cite{patrini2017making}            & 89.9    & - \\
        Distilling~\cite{zhang2020distilling}          & 92.7    & 90.2 \\
        Meta-Learning~\cite{MLNT}                         &  -      & 88.6 \\ 
        M-correction~\cite{arazo2019unsupervised}        &  -      & 86.3 \\ 
        Iterative-CV~\cite{chen2019understanding}        &  -      & 88.0 \\ 
        DivideMix~\cite{li2020dividemix}     & 93.4    & 93.4 \\ 
        REED ~\cite{zhang2020decoupling}                 & {95.0}    & {92.3}    \\
        C2D ~\cite{zheltonozhskii2022contrast}                 & {93.8}    & {93.4}    \\
        Sel-CL+ ~\cite{li2022selective}                 & 95.2    & {93.4}    \\
        GCE ~\cite{ghosh2021contrastive}                 & {87.3}    & {78.1}    \\
        RRL ~\cite{li2021learning}                 & -    & {92.4}    \\        
        \hline
        SANM(DivideMix)                                            & \textbf{95.4}    & \textbf{94.8}    \\
        \hline
    \end{tabular}}
    \label{tab:cifar10asy}
    
\end{minipage}
\hspace{1mm}
\begin{minipage}{0.53\linewidth}
    \centering
    \caption{Testing accuracy on Clothing-1M. 
    } 
     \vspace{-1em}
    \centering
    \footnotesize
    \renewcommand{\arraystretch}{1.15}
    {\begin{tabular}{l|c}
    \hline        
        Method                                       &  Acc   \\ \hline
        CrossEntropy                                 &  69.21                 \\ 
        F-correction~\cite{patrini2017making}        &  69.84                 \\ 
        M-correction~\cite{arazo2019unsupervised}    &  71.00                 \\ 
        Joint-Optim~\cite{tanaka2018joint}           &  72.16                 \\ 
        Meta-Cleaner~\cite{zhang2019metacleaner}     &  72.50                 \\ 
        Meta-Learning~\cite{MLNT}                    &  73.47                 \\ 
        PENCIL~\cite{yi2019probabilistic}            &  73.49                 \\ 
        Self-Learning~\cite{han2019deep}             &  74.45                  \\
        DivideMix~\cite{li2020dividemix}             &  74.76                  \\ 
        Nested~\cite{Chen_2021_CVPR}   & 74.90  \\
        AugDesc~\cite{nishi2021augmentation} & 75.11 \\
        RRL~\cite{li2021learning}  &  74.90            \\ 
        GCE~\cite{ghosh2021contrastive}  &  73.30            \\ 
        C2D~\cite{zheltonozhskii2022contrast}  &  74.30             \\ \hline
        SANM(DivideMix)                                     &  \textbf{75.63}   \\     
        \hline
    \end{tabular}}
    \label{tab:clothing1M}
    
\end{minipage}
\end{minipage}
\hspace{2mm}
\begin{minipage}{0.36\linewidth}
\begin{minipage}{0.9\linewidth}
    \caption{
        Comparison with masks generated from pre-trained backbones on CIFAR-10/100. M: Method. P: Pretrained Backbone. S: SANM. D: DivideMix. C: C2D.}
         \vspace{-1em}
    \renewcommand{\arraystretch}{1.1}
    \resizebox{0.98\linewidth}{!}{\begin{tabular}{l c  c |c  c  c c}
        \hline
            \multicolumn{3}{c|}{Dataset}     &    \multicolumn{4}{c}{CIFAR-10}                    \\  
        M & P &S  & 20\%       &       50\%       &     80\%      &     90\%      \\ \hline
        \multirow{2}{*}{D} & \Checkmark& \XSolidBrush& 96.0  &95.1  & 93.7 & 81.5   \\                                &\XSolidBrush &\Checkmark & \textbf{96.4} & \textbf{95.8} & \textbf{94.6} & \textbf{92.3}   \\  \hline
        \multirow{2}{*}{C} & \Checkmark& \XSolidBrush & 96.3 & 95.4  & 95.0   & 93.9  \\                               &\XSolidBrush &\Checkmark  & \textbf{96.6} & \textbf{96.4} & \textbf{95.7} & \textbf{95.1}  \\  
                                                                     \hline
    \end{tabular}}

    \centering
    \resizebox{0.98\linewidth}{!}{\begin{tabular}{l c  c |c  c  c c}
        \hline
            \multicolumn{3}{c|}{Dataset}     &    \multicolumn{4}{c}{CIFAR-100}                    \\  
        M & P &S  & 20\%       &       50\%       &     80\%      &     90\%      \\ \hline
        \multirow{2}{*}{D} & \Checkmark& \XSolidBrush & 78.8 & 76.6  & 64.9 & 36.4  \\                   
                           &\XSolidBrush &\Checkmark      & \textbf{81.2} & \textbf{78.2} & \textbf{68.7} & \textbf{43.5}  \\  \hline
        \multirow{2}{*}{C} & \Checkmark& \XSolidBrush &  80.0   & 77.1  & 69.2 & 58.4   \\                            
           &\XSolidBrush &\Checkmark & \textbf{81.9}   & \textbf{79.3} & \textbf{71.6} & \textbf{61.9}  \\  
                                                                     \hline
    \end{tabular}}
    \label{tab:cifar_perfect}
\end{minipage}
\end{minipage}
\vspace{-4mm}
 \end{table*}

\subsection{Experimental Results}
\vspace{-.5em}
$\bullet$ \textbf{Results on Simulated Noisy Datasets.} Here we compare the proposed method with multiple baselines on CIFAR-10/100 using different symmetric noise levels. As shown in Table.~\ref{tab:cifar_sym}, by integrating with C2D~\cite{zheltonozhskii2022contrast}, our SANM outperforms the naive C2D significantly and achieves state-of-the-art performance among these methods. Specifically, an increase of around $3\sim4$\% across all the symmetric noise cases on CIFAR-100 can be observed. Besides, when combined with DivideMix~\cite{li2020dividemix}, SANM can bring greater improvement to its performance, especially for severe label noise cases. Under 90\% symmetric noise setting, our SANM (DivideMix) outperforms DivideMix by a large margin of 12\%. A similar performance boost can be observed on the CIFAR-100 as well, especially in the high noise case. When compared with the implicit regularization method AugDesc~\cite{nishi2021augmentation}, our SANM (DivideMix) can also achieve much better accuracy across all the settings. Table.~\ref{tab:cifar10asy} shows comparisons of the test accuracy on CIFAR-10 with varied asymmetric noise levels. SANM shows a consistent improvement with respect to the state-of-the-art methods, demonstrating the ability of SANM in enhancing the robustness of the model to semantic noise.

\begin{table*}[!t]
\footnotesize
    \caption{
        Ablation study for the effectiveness of each key component. {AMG: adversarial noisy masking generation, NLR: noisy label regularization, SMR: self-supervised masking reconstruction}.} 
    \centering 
    \vspace{-1em}
    
    \begin{tabular}{c c c c |cccc|cccc}
    \hline
        \multicolumn{4}{c|}{Component}    &\multicolumn{4} {c|}{CIFAR-10} & \multicolumn{4}{c}{CIFAR-100} \\ 
                                          AMG & NLR & SMR   &     & 20\% & 50\% & 80\% & 90\%     & 20\% & 50\% & 80\% & 90\%      \\ \hline
        \multirow{2}{*}{\XSolidBrush} & \multirow{2}{*}{\XSolidBrush} & \multirow{2}{*}{\XSolidBrush}            & Best &96.1    & 94.6   & 93.2  & 76.0  & 77.3   &74.6    & 60.2  & 31.5   \\
                                    &                             &                            & Last &95.7    & 94.4   & 92.9  & 75.4  & 76.9   &74.2    & 59.6  & 31.0   \\ \hline
        \multirow{2}{*}{\Checkmark} & \multirow{2}{*}{\XSolidBrush} & \multirow{2}{*}{\XSolidBrush}        & Best &96.3    & 95.3   & 94.0  & 91.4  & 80.2   &77.3    & 68.0  & 42.7    \\
                                    &                             &                            & Last &96.2    & 95.1   & 93.6  & 90.8  & 79.7   &77.0    & 67.5  & 42.5   \\ \hline
        \multirow{2}{*}{\Checkmark} & \multirow{2}{*}{\Checkmark} & \multirow{2}{*}{\XSolidBrush}    & Best &96.4    & 95.5   & 94.2  & 91.6  & 80.5   &77.5    & 68.3  & 43.0   \\
                                    &                             &                            & Last &96.3    & 95.4   & 94.0  & 91.5  & 80.2   &77.1    & 68.2  & 42.7   \\ \hline
        \multirow{2}{*}{\Checkmark} & \multirow{2}{*}{\Checkmark} & \multirow{2}{*}{\Checkmark}& Best & \textbf{96.4} & \textbf{95.8} & \textbf{94.6} & \textbf{92.3} & \textbf{81.2} & \textbf{78.2} & \textbf{68.7} & \textbf{43.5} \\
                                    &                             &                            & Last & \textbf{96.3} & \textbf{95.6} & \textbf{94.3} & \textbf{92.1} & \textbf{80.4} & \textbf{78.0}& \textbf{68.3} & \textbf{43.0}\\ 
    \hline 
    \end{tabular} 
    
    \label{tab:ablation_backbone}
    \vspace{-2mm}
\end{table*}

\begin{table*}[!t]
\footnotesize
    \centering
    \caption{
    Comparison between the LNL methods and their SANM applications with symmetric noise on CIFAR-10/100. Specifically, the 9-layer CNN is adopted as the backbone network of Co-teaching.}
    \vspace{-1em}
    \setlength\tabcolsep{10pt}
    \resizebox{0.76\textwidth}{!}{$
    \begin{tabular}{l l |c|c|c|c|c|c|c|c}
        \toprule
        Dataset      &      &  \multicolumn{4}{c|}{CIFAR-10}       &  \multicolumn{4}{c}{CIFAR-100}            \\ 
        Method/Noise ratio &      & 20\% & 50\% &  80\%  & 90\%  & 20\% & 50\% &  80\%  & 90\%      \\ \midrule
        
        {CE}  & Best & 86.8&79.4&62.9&42.7&62.0&46.7&19.9&10.1   \\  \midrule
        {SANM(CE)}            & Best & \textbf{92.4}  & \textbf{89.7} & \textbf{72.1} & \textbf{51.5}  &  \textbf{70.9} & \textbf{53.1}  & \textbf{34.8}  & \textbf{18.6}\\ \midrule\midrule 
        {Co-teaching~\cite{Co-han2018co}}  & Best & 82.6  & 73.0 & 24.0 & 14.6  & 50.5 & 38.2  & 11.8  & 4.9  \\ \midrule
        {SANM(Co-teaching)}            & Best & \textbf{89.2}  & \textbf{78.2} & \textbf{36.4} & \textbf{20.7}  &  \textbf{58.2} & \textbf{51.3}  & \textbf{19.4}  & \textbf{13.4}\\   \midrule\midrule      
        {CDR~\cite{2021CDR}}  & Best &90.4   &85.0    &47.2   & 12.3 & 63.3   & 39.5   &29.2   & 8.0   \\ \midrule
        {SANM(CDR)}            & Best & \textbf{92.6}  & \textbf{91.6} & \textbf{55.3} & \textbf{16.7}  &  \textbf{72.7} & \textbf{56.4}  & \textbf{36.6}  & \textbf{20.8}\\  \midrule\midrule
        {ELR+~\cite{2020ELR}}    & Best & 94.6 & 93.8 & 91.1 &  75.2 & 77.5 & 72.4 & 58.2 & 30.8 \\  \midrule
        {SANM(ELR+)}              & Best & \textbf{96.3}  & \textbf{95.7} & \textbf{94.1} & \textbf{82.9}  &  \textbf{79.8} & \textbf{77.3}  & \textbf{65.0}  & \textbf{38.7}\\ 
                                             \bottomrule
    \end{tabular}$}
    \label{tab:cifar10_sym_combine}
    \vspace{-4mm}
\end{table*}

$\bullet$ \textbf{Results on Real-world Noisy Datasets.} 
The results on Clothing1M dataset are shown in Table.~\ref{tab:clothing1M}, one can see that SANM can bring around 0.87\% improvement in accuracy to the baseline method DivideMix, and get much better result than other state-of-the-art methods, verifying that SANM is also effective in real-world noise scenarios. 
Moreover, SANM obtains 89.3\% accuracy on Animal-10N (detailed comparison on Animal-10N can be found in our supplementary materials), which sets the new state-of-the-art, indicating that the proposed method can handle fine-grained real-world noisy datasets as well. 

\textbf{Generality \zbs{of SANM}.} 
To \zbs{further} verify the effectiveness and generalization of SANM, we \zbs{adopt vanilla} CE, Co-teaching~\cite{Co-han2018co}, CDR~\cite{2021CDR} and ELR+~~\cite{2020ELR}, other than DivideMix and C2D, \zbs{to work collaboratively with SANM}. As shown in Table.~\ref{tab:cifar10_sym_combine}, SANM brings consistent performance boost to all the corresponding baselines across all the noisy cases. For example, when seamlessly combined with SANM, an accuracy increase of around 14.9\% on the symmetric 80\% \zbs{noise setting (CIFAR-100)} can be achieved by vanilla CE, demonstrating that SANM is capable of effectively imposing adaptive regularization on noisy and clean samples respectively and thus alleviates over-fitting to noisy labels. Therefore, all the results indicate the generalization ability of SANM to boost existing LNL approaches. 

\vspace{-2mm}
\subsection{Ablation Study}

\textbf{Component Analysis.} To explore the impact of each component in SANM, i.e., AMG-adversarial noisy masking generation, NLR-noisy label regularization, SMR-self-supervised masking reconstruction, we conduct an ablation study on CIFAR-10 and CIFAR-100. As shown in Table.~\ref{tab:ablation_backbone}, we report test accuracy for all the noisy settings for SANM where the contribution of each component is clearly shown. Specifically, it can be observed that AMG is crucial to performance, especially under severe label noisy cases. Besides, the accuracy is {further} improved when further combined with NLR. By introducing noise-free self-supervised signals, SMR can also boost the performance of the baseline across all the settings. Best accuracy can be achieved by SANM when all the components are included. {This verifies the effectiveness of all three key designs in SANM.}

\textbf{Parameter Sensitivity.} {The basic mask ratio $\mu$ in Eq.~(\ref{eq:ratio}) is a important parameter within the proposed method. To investigate how $\mu$ affects the performance,} we conduct parameter analysis experiments under 50\%/80\% symmetric noise levels on CIFAR-10/100. Best test accuracy with respect to different $\mu$ is shown in Fig. \ref{fig:ablation_parameter}. Compared with the results of {baseline methods, i.e.,} masking the images with fixed ratios and random ratios that are randomly sampled from $Uniform(0,\mu)$ for each sample, SANM can achieve superior performance by adaptively generating instance-specific ratios. Moreover, the results show that keeping $\mu$ between 0.1 and 0.3 leads to better performance than without adversarial noisy masking ($\mu=0$). This is mainly because large $\mu$ makes the masked training images and the test images differ significantly in image style, and thus the model is unable to learn class-specific discriminative information when trained with such strong regularization. However, an extremely small $\mu$ {results in}
encoder overfit to noisy labels more severely. {Therefore}, we set $\mu$ as 0.2 to obtain the best performance on CIFAR-10/100. 

\textbf{Comparison with Masks from Pre-trained Backbones.} The effectiveness of the proposed adversarial masking generation is evaluated in Table.~\ref{tab:ablation_backbone}. And further we conduct experiments that utilizing a strong backbone (i.e., pre-trained ResNet-18 with DivideMix and C2D) to generate more object-centric masks as comparison to our method. The results shown in Table~\ref{tab:cifar_perfect} indicate that even compared with the mask generation strategy by well-trained strong backbones, our adversarial masking generation can consistently performs better since the mask generation strategy of fixed well-trained backbones cannot adaptively adjust the mask strategy according to the training state of the base model, leading to sub-optimal results.

\vspace{-2mm}
\section{Conclusion}
\vspace{-2mm}
In this paper, we propose SANM, a novel self-supervised adversarial noisy masking LNL method. SANM prevents models from overfitting noisy instances by explicitly imposing regularization on the features. A label quality guided adversarial masking scheme is designed to adaptively modify the inputs and noisy labels. Moreover, an auxiliary decoder is introduced to reconstruct images based on the modulated features, which enhances the generalization by providing extra noise-free self-supervised signals. Besides, SANM is flexible and can be easily plugged into existing LNL frameworks. State-of-the-art performance is obtained by SANM on both synthetic and real-world noisy datasets.

\medskip

{\small
\bibliographystyle{ieee_fullname}
\bibliography{egbib}
}

\end{document}